\documentclass[runningheads]{llncs}

\usepackage{eccv}

\usepackage{eccvabbrv}

\usepackage{url}
\usepackage{graphicx}
\usepackage{amsmath}
\usepackage{amssymb}
\usepackage{adjustbox}
\usepackage{color, colortbl}

\usepackage{booktabs}       
\usepackage{amsfonts}       
\usepackage{amsmath}
\usepackage{nicefrac}       
\usepackage{microtype}     
\usepackage{xcolor}        
\usepackage{multirow}
\usepackage{multicol}
\usepackage{xspace}
\usepackage{adjustbox}
\usepackage{import}
\usepackage{color, colortbl}
\usepackage{makecell}
\usepackage{wrapfig}
\usepackage{pifont}
\usepackage{sidecap}
\usepackage{enumitem}
\usepackage{threeparttable}

\usepackage{color}
\usepackage{xcolor}
\definecolor{citecolor}{HTML}{0071bc}
\definecolor{mlpMixerColor}{RGB}{242,122,130}
\definecolor{concatColor}{RGB}{0,118,186}
\definecolor{dwconvColor}{RGB}{254,174,0}

\usepackage[accsupp]{axessibility}  %

\usepackage[breaklinks,colorlinks,citecolor=eccvblue]{hyperref}

\usepackage{orcidlink}
\definecolor{VisGreen}{RGB}{1, 128, 0}

\newcolumntype{C}{>{\centering\arraybackslash}X}
\newcommand{\xmark}{\ding{55}}

\begin{document}

\title{EgoPoseFormer: A Simple Baseline for Stereo Egocentric 3D Human Pose Estimation} 
\titlerunning{EgoPoseFormer}

\author{Chenhongyi Yang\inst{1,2}\thanks{Work done when working as a research scientist intern at Meta Reality Labs} \and
Anastasia Tkach\inst{2} \and
Shreyas Hampali\inst{2} \and
Linguang Zhang\inst{2} \and
Elliot J. Crowley\inst{1} \and
Cem Keskin\inst{2}}

\authorrunning{C.~Yang et al.}

\institute{University of Edinburgh \and
Meta Reality Labs}

\maketitle

\newcommand{\ourmethod}{EgoPoseFormer}
\newcommand{\ourmethodFull}{Ego-centric Pose Estimation Transformer}

\newcommand{\ppn}{PPN}
\newcommand{\ppnFull}{Pose Proposal Network}

\newcommand{\rFormer}{PRFormer}
\newcommand{\rFormerFull}{Pose Refinement Transformer}

\newcommand{\dsattn}{Deformable Stereo Attention}

\begin{abstract}

We present \ourmethod, a simple yet effective transformer-based model for stereo egocentric human pose estimation. The main challenge in egocentric pose estimation is overcoming joint invisibility, which is caused by self-occlusion or a limited field of view (FOV) of head-mounted cameras. Our approach overcomes this challenge by incorporating a two-stage pose estimation paradigm: in the first stage, our model leverages the global information to estimate each joint’s coarse location, then in the second stage, it employs a DETR style transformer to refine the coarse locations by exploiting fine-grained stereo visual features. In addition, we present a \dsattn~operation to enable our transformer to effectively process multi-view features, which enables it to accurately localize each joint in the 3D world. We evaluate our method on the stereo UnrealEgo dataset and show it significantly outperforms previous approaches while being computationally efficient: it improves MPJPE by 27.4mm (45\% improvement) with only 7.9\% model parameters and 13.1\% FLOPs compared to the state-of-the-art. Surprisingly, with proper training settings, we find that even our first-stage pose proposal network can achieve superior performance compared to previous arts. We also show that our method can be seamlessly extended to monocular settings, which achieves state-of-the-art performance on the SceneEgo dataset, improving MPJPE by 25.5mm (21\% improvement) compared to the best existing method with only 60.7\% model parameters and 36.4\% FLOPs. Code is available at \url{https://github.com/ChenhongyiYang/egoposeformer}.

\end{abstract}

\section{Introduction}
\label{sec:intro}

In the rapid expansion of Virtual Reality (VR) and Augmented Reality (AR) technologies~\cite{applevisionpro,metaquest}, the capability to accurately interpret and emulate human actions becomes increasingly crucial. Central to this pursuit is the egocentric pose estimation task~\cite{jiang2021egocentric, tome2020selfpose, kang2023ego3dpose, wang2023scene, wang2021estimating, hakada2022unrealego, tome2019xr, wang2022estimating, rhodin2016egocap, xu2019mo, park2022building, zhao2021egoglass, li2023ego}, which aims to estimate the 3D body pose from a vantage point inherent to the user, predominantly from head-mounted cameras. Its precision is pivotal in a wide range of applications, such as gaming and virtual meetings, making it essential for crafting an immersive user experience for the next generation of VR/AR systems.

Different from outside-in pose estimation, where the human bodies are well covered by the images, a key challenge of egocentric pose estimation is the joint invisibility problem, which usually results from two causes. First, the limited field of view (FOV) of head-mounted cameras cannot fully capture the human body~\cite{hakada2022unrealego,tome2019xr}, especially when hands and legs are stretched out. Another cause arises from the self-occlusion of different body parts~\cite{tome2019xr}, especially the lower body, which is very prone to be occluded by the main trunk. To overcome this limitation, some recent works~\cite{tome2019xr,hakada2022unrealego,zhao2021egoglass} directly regress the 3D joint locations using the 2D heatmaps by employing an auto-encoder style architecture, which allows the locations of invisible joints to be inferred from the global information and other visible joints' locations. 
These approaches, however, take 2D heatmaps as input and cannot leverage the rich appearance information of the input image, which limits its 3D regression capability.
This also causes a poor scaling-up ability to the developed model, i.e., even if the model is equipped with a larger backbone network, its pose estimation accuracy cannot be improved~\cite{hakada2022unrealego}. In another recent work, SceneEgo~\cite{wang2023scene}, a 3D feature voxel grid was first built using fish-eye projection~\cite{scaramuzza2006flexible} with the help of depth and semantic maps, and 3D convolution was employed to operate on the feature grid to estimate each joint's location in a monocular setting. 
Despite the fact that 3D joints outside the FOV can be estimated by leveraging a sufficiently large voxel grid, 3D convolutions are computationally expensive and the cost increases with the size of the voxel grid.

\begin{figure}[!t]
\centering
    \includegraphics[width=\linewidth]{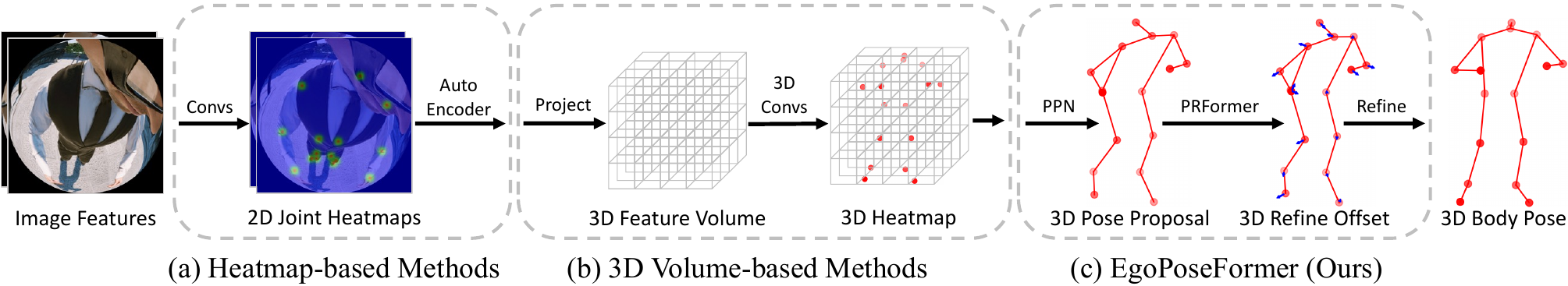}
    \caption{\footnotesize Illustration of different egocentric pose estimation methods. While previous approaches predict joints' locations via 2D heatmaps or 3D feature voxels, \ourmethod~first estimates the coarse locations of each joint using a \ppnFull~(\ppn) and uses a transformer to refine the estimated pose.}
    \label{fig:teaser}
\end{figure}

In this work, we propose ~\emph{\ourmethod}, a simple transformer-based model for multi-view egocentric pose estimation. As illustrated in~\cref{fig:teaser}, our method uses a two-stage framework~\cite{ren2015faster} to overcome the joint invisibility challenge, and the body pose is predicted in a coarse-to-fine manner~\cite{Wang_2023_ICCV,ren2023decoupled,wang2023deep}. Specifically, the first stage of our model is the \ppnFull~(\ppn), which is a simple 2-layer MLP that leverages the global information of the input multi-view feature maps to predict a joint's coarse location. Similar to the recent heatmap-based methods~\cite{hakada2022unrealego,tome2019xr}, the usage of global features allows our method to reason about the locations of all joints, including the invisible ones. Surprisingly, we found, with proper training settings, this simple MLP can already outperform previous state-of-the-art methods in stereo inputs are available. Then, in the second stage, we employed a DETR-style~\cite{carion2020end} transformer, \rFormerFull~(\rFormer), to predict 3D refinement offsets related to the first stage estimations by exploiting the multi-view stereo features and human kinematic information.  Specifically, we embed each joint's location and identity information into a Joint Query Token (JQT). Each JQT interacts with the multi-view image features and other JQTs through attention operations in each layer of \rFormer; subsequently, the refinement offsets are predicted from updated JQTs. Furthermore, we design a new \dsattn~\cite{zhu2021deformable} to effectively process the fine-grained multi-view stereo features, which allows us to accurately estimate a joint's 3D location. In summary, we make three  contributions: 
\begin{itemize}
    \item We propose \ourmethod, a simple transformer-based model for stereo egocentric pose estimation. Our model composes an MLP-based pose proposal network for computing coarse joint locations, which already demonstrates a strong accuracy, and a transformer-based pose refinement network to further improve the localization accuracy. 
    \item Our method achieves state-of-the-art on the stereo UnrealEgo dataset~\cite{hakada2022unrealego} by a huge advantage over previous arts with much lower computation costs. 
    \item We demonstrate that our method can be easily extended to the monocular egocentric pose estimation problem and achieve state-of-the-art performance on the SceneEgo dataset~\cite{wang2023scene}. 
\end{itemize}

\section{Related Work}
\label{sec:related}

\noindent\textbf{Egocentric Pose Estimation.}
Prior to our work, there have been several works on egocentric pose estimation, which cover both monocular and stereo settings. Most previous approaches are based on predicting 3D joints locations from 2D heatmaps. For example, Mo$^2$Cap$^2$~\cite{xu2019mo} first predicts the 2D joint heatmaps and their corresponding depth, and the 3D coordinates are computed with fish-eye unprojection. In xR-EgoPose~\cite{tome2019xr}, the 3D joint coordinates are directly estimated with an auto-encoder, whose input is the predicted joint heatmap. This allows it to tackle the joint invisibility difficulty. SelfPose~\cite{tome2020selfpose} improves xR-EgoPose by introducing joint rotation loss and UNet~\cite{ronneberger2015u} backbone. EgoSTAN~\cite{park2022building} improves the quality of visual features by introducing temporal modeling. EgoGlass~\cite{zhao2021egoglass} extends the monocular heatmap-based methods to multi-view settings, where the joints' locations are estimated from the multi-view joint heatmaps. It also adds an auxiliary segmentation loss to improve accuracy further. EgoPW~\cite{wang2021estimating} explores extending existing pose estimation methods to estimate body poses in a global space. UnrealEgo~\cite{hakada2022unrealego} further improves EgoGlass by introducing cross-view information exchange in the UNet decoder. The recently proposed Ego3DPose~\cite{kang2023ego3dpose} improves UnrealEgo by explicitly modeling limb heatmaps and orientations. Apart from heatmap-based approaches, SceneEgo~\cite{wang2023scene} was recently introduced to directly predict joint locations by running 3D convolution on 3D feature voxels with the help of scene depth and segmentation. There are also works~\cite{li2023ego, luo2021dynamics,yuan20183d} about egocentric pose hallucination, where the headset wearer's body pose is estimated with front-facing cameras, in which the body is rarely observed. Different from ours, the focus of those works is generating body poses that are harmonious with the background scene. \\

\begin{figure*}[!t]
    \centering
    \includegraphics[width=1.0\textwidth]{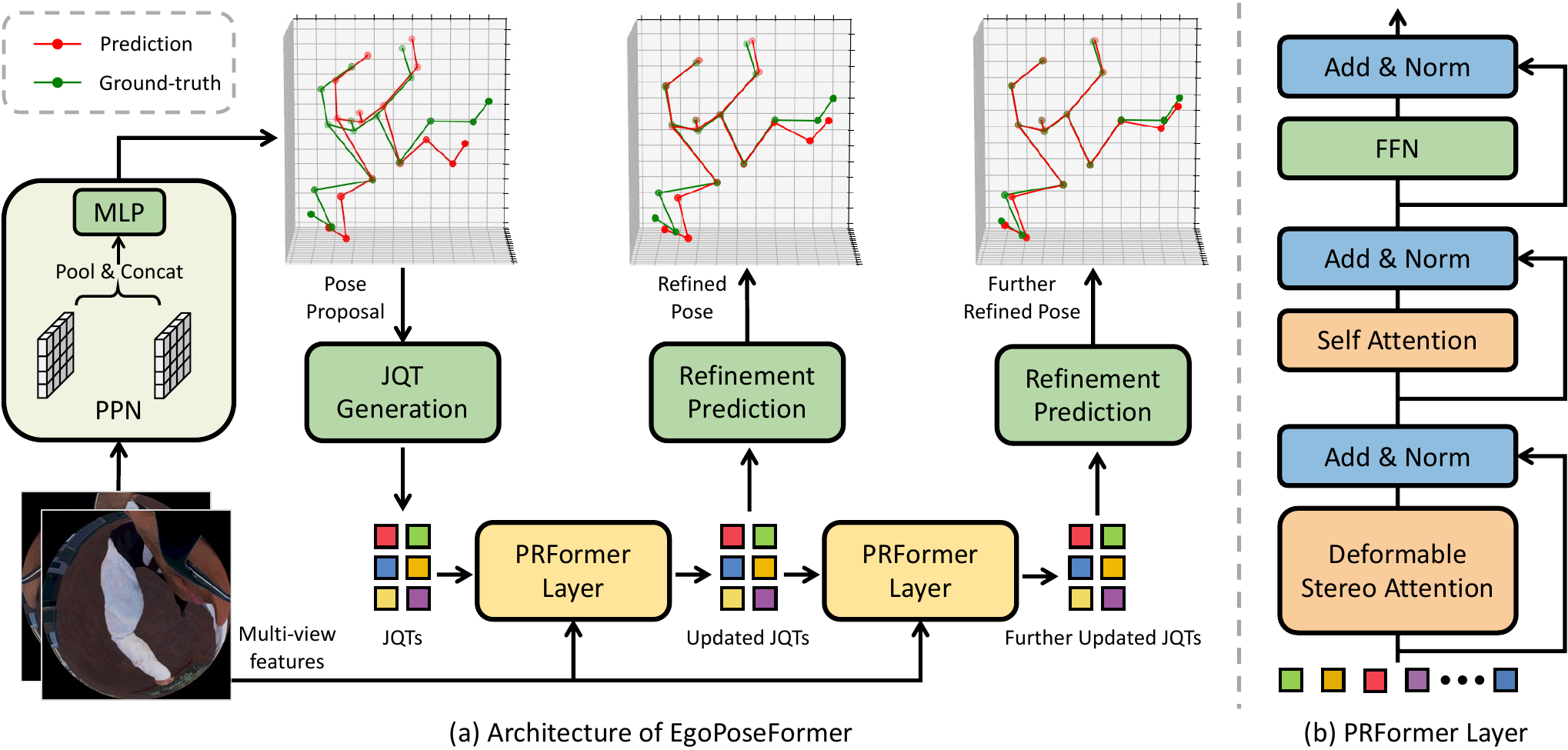}
    \caption{\footnotesize \textbf{(a)} An overview of the proposed \ourmethod. The input of \ourmethod~is the multi-view image features. In the first stage: \ppnFull~(\ppn), the multi-view features are globally pooled and concatenated, from which an MLP is used to estimate the coarse location of each joint (pose proposal). Then the joints' identity and location information are embedded into Joint Query Tokens (JQTs) to serve as the queries in the second stage \rFormerFull~(\rFormer). In \rFormer, a JQT iteratively interacts with the stereo features and other JQTs to update itself through the attention mechanism. The updated JQTs are used to predict refinement offsets related to the pose proposal, yielding more accurate pose estimations. \textbf{(b)} The architecture of \rFormer~layer is similar to the transformer decoder layer, which includes a cross-attention block, a self-attention block, and a feed-forward network (FFN). However, in \rFormer, the cross-attention is replaced by the proposed \dsattn~to better exploit stereo visual features.}
    \label{fig:network}
\end{figure*}

\noindent\textbf{Transformer for Outside-in Pose Estimation.}
There have been lots of successful attempts to apply transformers for the outside-in body pose estimation.
One line of work~\cite{zhang2022mixste, zhao2023poseformerv2, zheng20213d, li2022mhformer, shan2022p, einfalt2023uplift} aims to develop high-performing transformer-based backbone networks for outside-in pose estimation. For example, PoseFormer~\cite{zheng20213d} introduces spatial and temporal attention mechanisms to generate high-quality visual features for 3D pose estimation, which is improved by the followed-up PoseFormer v2~\cite{zhao2023poseformerv2} by introducing frequency modelling. Furthermore, improvements in model efficiency are also elaborated in~\cite{einfalt2023uplift, li2022exploiting}. The other line of work, similar to ours, focuses on developing DETR-style~\cite{liu2023group, shi2022end, qiu2023psvt, choi2022learning, xiao2022querypose, yang2023explicit} sparse transformers for human body pose estimation. For example, PETR~\cite{shi2022end} introduces inter-instance and intra-instance attention for accurate 2D multi-person pose estimation. PSVT~\cite{qiu2023psvt} introduces spatial-temporal encoder and decoder for 3D pose and body shape estimation. On the other hand, transformer architecture is also used for 2D human pose estimation~\cite{xiao2022querypose,liu2023GroupPose}. For example, GroupPose~\cite{liu2023GroupPose} uses keypoint and instance queries to directly estimate the 2D human poses in a multi-person setting. Despite the huge success of transformer architectures in outside-in pose estimation, applying it to egocentric settings requires non-trivial adaptation due of the intrinsic difference between the two problems. For instance, in outside-in pose estimation, the human body usually lies within the camera's FOV, while the out-of-FOV problems usually happen in egocentric pose estimation. Another difference is the input of those two tasks: most outside-in pose estimation models take regular images captured by pin-hole cameras as input~\cite{ionescu2013human3,lin2014microsoft}, while in egocentric settings the images are usually captured by fish-eye cameras~\cite{hakada2022unrealego,zhao2021egoglass,wang2023scene} to expand FOV, causing image distortions and posing further difficulties to the task.

\section{Method}
\label{sec:method}

In this section, we introduce the proposed \ourmethod~in detail. We first discuss the motivation of our two-stage framework in~\cref{sec:twoStage}. We present the \ppnFull~in\cref{sec:ppn} and the \rFormerFull~in~\cref{sec:former}. Later, we introduce the loss function in~\cref{sec:loss} and the feature extractor in~\cref{sec:encoder}. An overview of our model is illustrated in~\cref{fig:network} (a).

\subsection{Two-stage Pose Estimator}
\label{sec:twoStage}

\cref{fig:network} (a) shows the proposed two-stage framework, which is designed to overcome the joint invisibility challenge caused by self-occlusion or the limited FOV of head-mounted cameras.
In the first stage, we estimate the coarse location of each joint, which we call \textit{pose proposal}~\cite{ren2015faster}, by utilizing the global feature pooled from the stereo features. This design enables the network to roughly localize all joints, including the invisible ones, by jointly reasoning visual clues from visible joints and background scenes. 
The global feature, however, does not preserve fine-grained local details that enable more accurate joint localization.
Motivated by this observation, we added a second stage to refine the pose proposal, where we use a transformer that exploits fine-grained stereo features and the body kinematic information through the attention mechanism. Finally, the refined poses are output as the final pose estimation results. 

\subsection{\ppnFull~(\ppn)}
\label{sec:ppn}

Given the multi-view image features $\mathbf{F} \in \mathbb{R}^{V \times H \times W \times C}$, the \ppn~computes the pose proposal $P^0 \in \mathbb{R}^{N_j \times 3}$ using global information. Here $V$ denotes the number of views; $H$, $W$ and $C$ denotes the height, width and channel number of the image feature maps; $N_j$ denotes the number of body joints. Specifically, \ppn~begins by applying a global average pooling to the feature maps of each view, then the resultant averaged features are concatenated to form a unified feature representation, which captures the salient global features and stereo information across views. We then predict each joint's initial 3D location with a 2-layer MLP with GELU activation: 
\begin{align}
P^0 = \mathrm{MLP}_{\mathrm{ppn}}\Bigl(\mathrm{Concat}_{\{k\}}\bigl(\mathrm{AvgPool}(F_k)\bigr)\Bigr)
\end{align}
where $k \in \{1, ..., V\}$. As we will show in~\cref{exp:ablation}, this simple design can already provide reasonable location estimation for each joint.

\subsection{\rFormerFull~(\rFormer)}
\label{sec:former}

The \rFormer~takes the multi-view image features $\mathbf{F} \in \mathbb{R}^{V \times H \times W \times C}$ and the pose proposal $P^0 \in \mathbb{R}^{N_j \times 3}$ as inputs. Structurally, the transformer comprises $S$ layers. At each layer $s$, a refinement offset $\Delta P^s \in \mathbb{R}^{N_j \times 3}$ relative to $P^0$ is predicted, and the pose estimation is computed by adding the offset to the pose proposals $P^s=\Delta P^s + P^0$ where $s=\{1, ..., S\}$. During the inference phase, the final layer's output, $P^S$, is used as the model's final prediction. \\

\begin{figure*}[!t]
    \centering
    \includegraphics[width=1.0\textwidth]{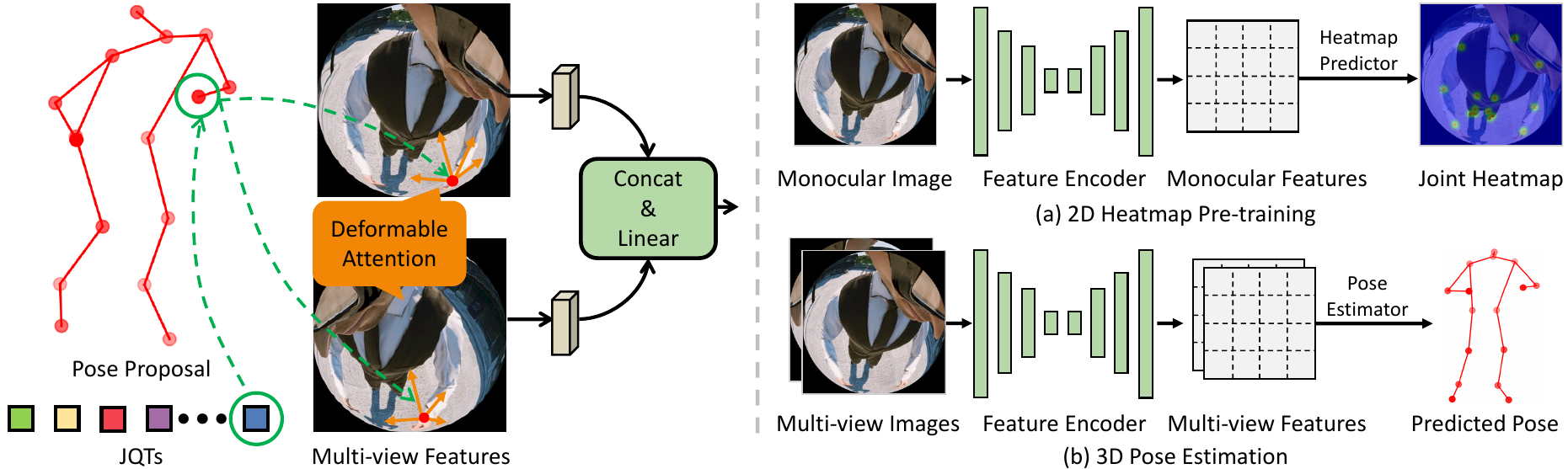}
    \caption{\footnotesize \textbf{Left}: An illustration of our~\dsattn. The 3D joints are first projected to each view plane using camera parameters. Within each view, we compute 2D deformable attention by querying the image features with the JQTs with the projected points serving as reference points. Finally, the attention results for each view are concatenated and fed into a linear layer to be projected into the original dimension. \textbf{Right}: (a) The feature extractor is first pre-trained to predict 2D joint heatmaps using monocular images. (b) The multi-view feature maps are computed using the pre-trained feature extractor.}
    \label{fig:dattnPretrain}
\end{figure*}

\noindent\textbf{Joint Query Tokens.}
Inspired by DETR~\cite{carion2020end}, in \rFormer~every joint is characterized by a unique Joint Query Token (JQT). The JQTs will serve as the queries in our transformer to interact with each other and the multi-view image features through attention mechanisms. We compute the JQTs by embedding each joint's identity and initial location information with a Query Generation MLP. Specifically, for the $j$-th joint, its JQT $Q_j \in \mathbb{R}^{C}$ is computed by feeding its initial location $P^0_j=(x^0_j, y^0_j, z^0_j)$ and a scalar identifier $\sigma_j$ into the MLP:  
\begin{align}
    Q_j= \mathrm{MLP}_{\mathrm{JQT}}\bigl(\sigma_j, x^0_j, y^0_j, z^0_j\bigr)
\end{align}
In practice, we simply use the joint index $j$ to serve as the scalar identifier $\sigma_j$. \\

\noindent\textbf{\rFormer~Layer.}
As shown in~\cref{fig:network} (b), each layer of the \rFormer~is a transformer decoder layer~\cite{vaswani2017attention}. Each input JQT undergoes a cross-attention operation to interact with the fine-grained stereo image features, and a subsequent self-attention operation to extract spatial and human kinematic information from other JQTs. Note that here we follow~\cite{Cheng_2022_CVPR} to put the cross-attention operation before the self-attention operation. Finally, we use a Feed-forward Network (FFN)~\cite{vaswani2017attention} to non-linearly transform the JQTs. A distinct characteristic of the \rFormer~layer vis-à-vis traditional transformer decoder layers is our replacement of the conventional cross-attention with \dsattn, which enables our transformer to effectively reason about multi-view stereo and thus can accurately locate the joints in the 3D world. \\

\noindent\textbf{\dsattn.}
As shown in~\cref{fig:dattnPretrain} Left, the proposed \dsattn~has three steps. Firstly, we project each body joint's initial 3D location onto each view plane by leveraging the camera parameters~\cite{scaramuzza2006flexible}. Specifically, for joint $j$ and its initial 3D location $P^0_j$, we compute its 2D location on each view $\{\tilde{P}^k_j\}=\{(\tilde{x}^k_j, \tilde{y}^k_j)\}$ where $k=\{1, ..., V\}$. Secondly, with the joints' projected 2D locations as reference points, we independently apply deformable attention~\cite{zhu2021deformable} in each view to let the JQTs extract useful information from image features:
\begin{align}
    Z^k_j = \mathrm{DeformAttn}\bigl(Q_j, \tilde{P}^k_j, F_k\bigr)
\end{align}
where $Q_j$ is the $j$-th JQT and $F_k$ is the image feature of the $k$-th view. Notably, if a joint is out of the view's FOV, we simply fill the computed result $Z^k_j$ with zeros to make this information explicit to the model. Finally, the computed results $\{Z^k_j\}$ from each view are concatenated and fed into a linear projection layer to fuse multi-view stereo information and transform the result to the original dimension:
\begin{align}
    Z_j = \mathrm{Linear}\bigl( \mathrm{Concat}_{\{k\}}(Z^k_j)\bigr)
\end{align}
In this way, the \dsattn~becomes an \textit{atomic} attention operation and can replace the normal cross-attention operation in a plug-and-play manner. \\

\noindent\textbf{Predicting Refinement Offsets.}
Each \rFormer~layer empowers a JQT to engage with the multi-view image features and other JQTs to exploit stereo and human kinematic information, imbuing it with joint-specific knowledge. Based on this enriched representation, we use a shallow MLP to predict the refinement offset relative to the pose proposal. Specifically, the refinement offset for joint $j$ at the $s$-th \rFormer~layer is computed as:
\begin{align}
\Delta P^s_j= \mathrm{MLP}_{\mathrm{offset}}^s(Q^s_j)
\end{align}

\subsection{Loss Function}
\label{sec:loss}

To keep our method concise, we employ a simple per-joint error loss~\cite{hakada2022unrealego} to train our model. Specifically, the loss function is formulated as the following: 
\begin{align}
    \mathcal{L} = \sum_{s=0}^{S} \sum_{j=1}^{N_j}{||P^{s}_{j}-\hat{P}_j||_{2}}
\end{align}
Here, different loss stages are indexed with $s$. Specifically, $s=0$ corresponds to the pose proposals generated by \ppn, and $s>0$ represents the subsequent refinement predictions derived from the transformer layers.  $N_j$ is the total number of joints; $P^{*}_{j}$ and $\hat{P}_j$ are the predicted and ground-truth 3D coordinates of the $j$-th joint.

\subsection{Feature Extractor}
\label{sec:encoder}

We follow UnrealEgo~\cite{hakada2022unrealego} to adopt a UNet~\cite{ronneberger2015u} architecture as our visual feature extractor to compute the multi-view image features. The key difference in our method is the exclusion of multi-view concatenation in the decoder part~\cite{hakada2022unrealego}. As the \rFormer~can effectively process multi-view stereo features, we can extract features from each view independently instead of aggregating multi-view features within the feature extractor. As we show in~\cref{exp:main}, this modification significantly reduces the model size and improves its computational efficiency.

Inspired by the heatmap-based methods~\cite{tome2019xr,hakada2022unrealego}, we further enhance our model's efficacy by pre-training its feature extractor to predict 2D joint heatmaps. As shown in~\cref{fig:dattnPretrain} right, during the pre-training phase, the model predicts 2D joint heatmaps for each view using a light-weighted fully convolutional head~\cite{hakada2022unrealego}, which is removed after pre-training.

\section{Experiments}
\label{sec:exp}

\subsection{Experiment Settings}
\label{exp:settings}

\noindent\textbf{Dataset Settings.}
We use the multi-view UnrealEgo~\cite{hakada2022unrealego} dataset to benchmark our proposed method.  The UnrealEgo dataset has 451k synthetic stereo views collected using 30 different actions, which are captured by two head-mounted cameras placed 1cm from the head. We follow the official dataset splits: the model is trained using the training set (357k views) and evaluated using the test set (48k views); The validation set (46k views) is used for tuning the hyperparameters. We also test our method using the monocular SceneEgo~\cite{wang2023scene} dataset to test its generalization ability to monocular settings. SceneEgo is a real-human dataset recorded by two actors with different daily actions. It has a total 28k images. For both datsets, we report the Mean Per Joint Position Error (MPJPE) and Procrustes Analysis MPJPE (PA-MPJPE) in millimeters as evaluation metrics following their official papers~\cite{hakada2022unrealego,wang2023scene}. \\

\noindent\textbf{Model and Training Settings.}
Following their original papers~\cite{hakada2022unrealego,wang2023scene}, we use ResNet-18~\cite{he2016deep} as the backbone network for UnrealEgo and ResNet-50 for SceneEgo. Following~\cite{hakada2022unrealego}, we set the visual feature's down-sampling stride to 4. We employ three layers of transformer decoder in our \rFormer. We use very similar training recipes for both datasets without careful tuning. Specifically, for both 2D joint heatmap pre-training and 3D pose estimation, we set the batch size to 16 and trained the model for 12 epochs. AdamW~\cite{loshchilov2018decoupled} is used as the default optimizer. The initial learning rate is set to 0.001 and weight decay is set to 0.005. We set the gradient clip to 5.0. The learning rates are decayed by a factor of 10 after the 8th and 11th epochs. 
\begin{table}[t!]
  \centering
  \caption{\small Comparison of MPJPE (PA-MPJPE) between our method and previous state-of-the-art approaches on the UnrealEgo dataset.}
    \label{tab:unrealego}
  \resizebox{\textwidth}{!}{
  \begin{tabular}{l|cccccccccc|c}
    \toprule
    Method& {Jump}& {Fall}& {Exercise}& {Pull}& {Sing}& {Roll}& {Crawl}& {Lay}& {Sitting}& {Crouch-N}& \\
    \hline
    xr-EgoPose\cite{tome2019xr} & 106.3 ( - ) & 167.2 ( - ) & 133.2 ( - ) & 119.5 ( - ) & 99.6 ( - ) & 116.1 ( - ) & 223.5 ( - ) & 146.7 ( - ) & 274.9 ( - ) & 172.2 ( - ) &  \\
    EgoGlass\cite{zhao2021egoglass} & 81.3(63.4) & 131.6(94.7) & 100.2(74.1) & 81.9(62.0) & 70.6(52.6) & 103.3(92.7) & 182.4(113.6) & 109.6(81.7) & 207.3(114.7) & 132.7(110.8) &  \\
    UnrealEgo\cite{hakada2022unrealego} & 76.6(61.1) & 126.8(95.9) & 90.3(68.4) & 78.2(61.4) & 67.3(49.8) & 86.2(71.6) & 181.5(116.4) & 97.8(76.6) & 194.3(150.3) & 116.8(97.9)  & \\
    Ego3DPose\cite{kang2023ego3dpose}  & 60.8(49.4) & 95.8(79.1) & 76.1(63.3) & 58.1(45.7) & 51.7(41.0) & 81.9(71.1) &148.7(105.1) & 83.4(69.8) & 153.8(133.3) & 93.6(79.6) & \\
    \hline
    Ours  & \textbf{34.9(35.2)} & \textbf{71.0(63.9)} & \textbf{40.0(39.6)} & \textbf{24.9(24.1)} & \textbf{27.2(25.8)} & \textbf{42.4(45.3)} & \textbf{99.5(77.4)} & \textbf{61.7(59.7)} & \textbf{98.4(99.3)} & \textbf{54.1(52.9)}  & \\
    \hline
    \hline
    Method& {Crouch-T} & {Crouch-TS}& {Crouch-F}& {Crouch-B}& {Crouch-S}& {Stand-WB}& {Stand-UB}& {Stand-T}& {Stand-TC}& {Stand-F} &  \\
    \hline
    xr-EgoPose\cite{tome2019xr}   & 173.8 ( - ) & 108.9 ( - ) & 119.9 ( - ) & 136.5 ( - ) & 145.8 ( - ) & 94.3 ( - ) & 93.3 ( - ) & 103.3 ( - ) & 101.6 ( - ) & 99.7 ( - )  & \\
    EgoGlass\cite{zhao2021egoglass}  & 128.5(110.8) & 82.3(60.5) & 79.2(65.4) & 94.8(77.9) & 93.9(77.9) & 70.2(51.5) & 70.2(46.3) & 77.7(59.3) & 77.8(70.0) & 77.3(62.3) & \\
    UnrealEgo\cite{hakada2022unrealego} & 128.2(104.3) & 76.8(55.2) & 73.6(61.9) & 78.2(62.0) & 84.9(71.4) & 68.5(50.1) & 66.5(45.4) & 74.3(57.9) & 74.1(61.2) & 70.8(56.6) & \\
    Ego3DPose\cite{kang2023ego3dpose} & 109.0(89.9) & 65.3(48.7) & 53.9(45.9) & 58.1(46.9) & 67.7(56.7) & 50.4(39.9) & 50.5(37.2) & 58.5(48.1) & 59.6(52.8) & 55.7(48.7) & \\

    \hline
    Ours & \textbf{78.4(69.3)} & \textbf{27.9(26.1)} & \textbf{23.5(25.3)} & \textbf{28.9(29.6)} & \textbf{33.1(36.5)} & \textbf{26.2(25.3)} & \textbf{25.6(24.3)} & \textbf{30.4(32.8)} & \textbf{22.2(23.2)} & \textbf{33.8(37.2)}  & \\
    \hline
    \hline

    Method & {Stand-B}& {Stand-S} & {Dance}& {Boxing}& {Wrestling}& {Soccer}& {Baseball}& {Basketball}& {Gridiron}& {Golf}& All \\ 
    \hline 
    xr-EgoPose\cite{tome2019xr} & 105.8 ( - ) & 114.3 ( - ) & 116.7 ( - ) & 97.3 ( - ) & 116.6 ( - ) & 104.6 ( - ) & 103.7 ( - ) & 98.6 ( - ) & 149.7 ( - ) & 117.5 ( - ) & 112.6 ( - ) \\
    EgoGlass\cite{zhao2021egoglass}  & 75.1(58.5) & 85.2(67.8) & 83.7(65.5) & 71.3(54.2) & 86.1(65.3) & 52.3(58.2) & 77.7(61.0) & 59.7(47.3) & 102.1(80.5) &69.4(48.2)  & 83.3(61.6) \\
    UnrealEgo\cite{hakada2022unrealego} & 68.1(55.1) & 78.9(62.3) & 79.9(63.2) & 68.7(51.4) & 83.7(64.5) & 78.3(56.2) & 72.8(56.7) & 60.5(45.4) & 99.5(81.8) & 73.8(49.0) & 79.1(59.2) \\
    Ego3DPose\cite{kang2023ego3dpose} & 51.9(42.6) & 60.9(51.2) & 61.3(50.8) & 49.9(40.2) &65.5(51.7) & 54.4(43.7) & 69.2(55.6) &48.4(35.8) &83.3(72.9) &54.5(38.1) & 60.8(48.5) \\
    \hline
    Ours & \textbf{26.8(28.5)} & \textbf{32.2(34.5)} & \textbf{35.0(36.2)} & \textbf{22.9(22.6)} &\textbf{37.0(36.6)} & \textbf{27.0(27.4)} & \textbf{41.6(42.9)} &\textbf{16.5(16.1)} &\textbf{58.4(48.7)} &\textbf{26.9(21.8)} & \textbf{33.4(32.7)} \\
    \bottomrule
  \end{tabular}}
\end{table}

\begin{table}[t]
    \begin{minipage}[b]{0.42\textwidth}
     \centering
    \caption{\footnotesize Comparison of MPJPE and PA-MPJPE with other methods on the monocular SceneEgo dataset.}
    \label{tbl:sceneEgo}
    \begin{adjustbox}{max width=\textwidth}
    \begin{tabular}{l@{\hskip 20pt}|cc}
        \toprule
    {Method}  &MPJPE & PA-MPJPE \\
    \midrule
    Mo$^{2}$Cap$^{2}$~\cite{xu2019mo} & 200.3 & 121.2 \\
    xR-egopose~\cite{tome2019xr} & 241.3 & 133.9 \\
    EgoPW~\cite{wang2021estimating} & 189.6 & 105.3 \\
    SceneEgo~\cite{wang2023scene} & 118.5 & 92.7 \\
    \midrule
    Ours & \textbf{93.0} & \textbf{74.3} \\
    \bottomrule
    \end{tabular}
    \end{adjustbox}
    \end{minipage}%
    \hfill
    \begin{minipage}[b]{0.545\textwidth}
    \centering
    \caption{\footnotesize Comparison of model efficiency between \ourmethod~and other methods on UnrealEgo and SceneEgo datasets.}
    \label{tbl:flops}
    \begin{adjustbox}{max width=\textwidth}
    \begin{tabular}{l|l|cc|rr}
    \toprule
     Dataset & Methods & MPJPE & PA-MPJPE & Params & FLOPs\\
    \midrule
     \multirow{4}{*}{UnrealEgo} & EgoGlass & 83.3 & 61.6 & 107.3 M & 16.1 G \\
      & UnrealEgo & 79.1 & 59.2 & 106.8 M & 27.1 G \\
      & Ego3DPose & 60.8 & 48.5 & 178.4 M & 55.6 G \\
      & Ours & \textbf{33.4} & \textbf{32.7} & \textbf{14.1} M & \textbf{7.3} G  \\
    \midrule
    \multirow{2}{*}{SceneEgo} & SceneEgo & 118.5 & 92.7 & 45.9 M & 157.3 G \\
        & Ours & \textbf{93.0} & \textbf{74.3} & \textbf{27.9} M & \textbf{50.4} G \\
    \bottomrule
    \end{tabular} 
    \end{adjustbox}
    \end{minipage}
\end{table}

\subsection{Main Results}
\label{exp:main}

\begin{figure*}[!t]
    \centering
    \includegraphics[width=1.0\textwidth]{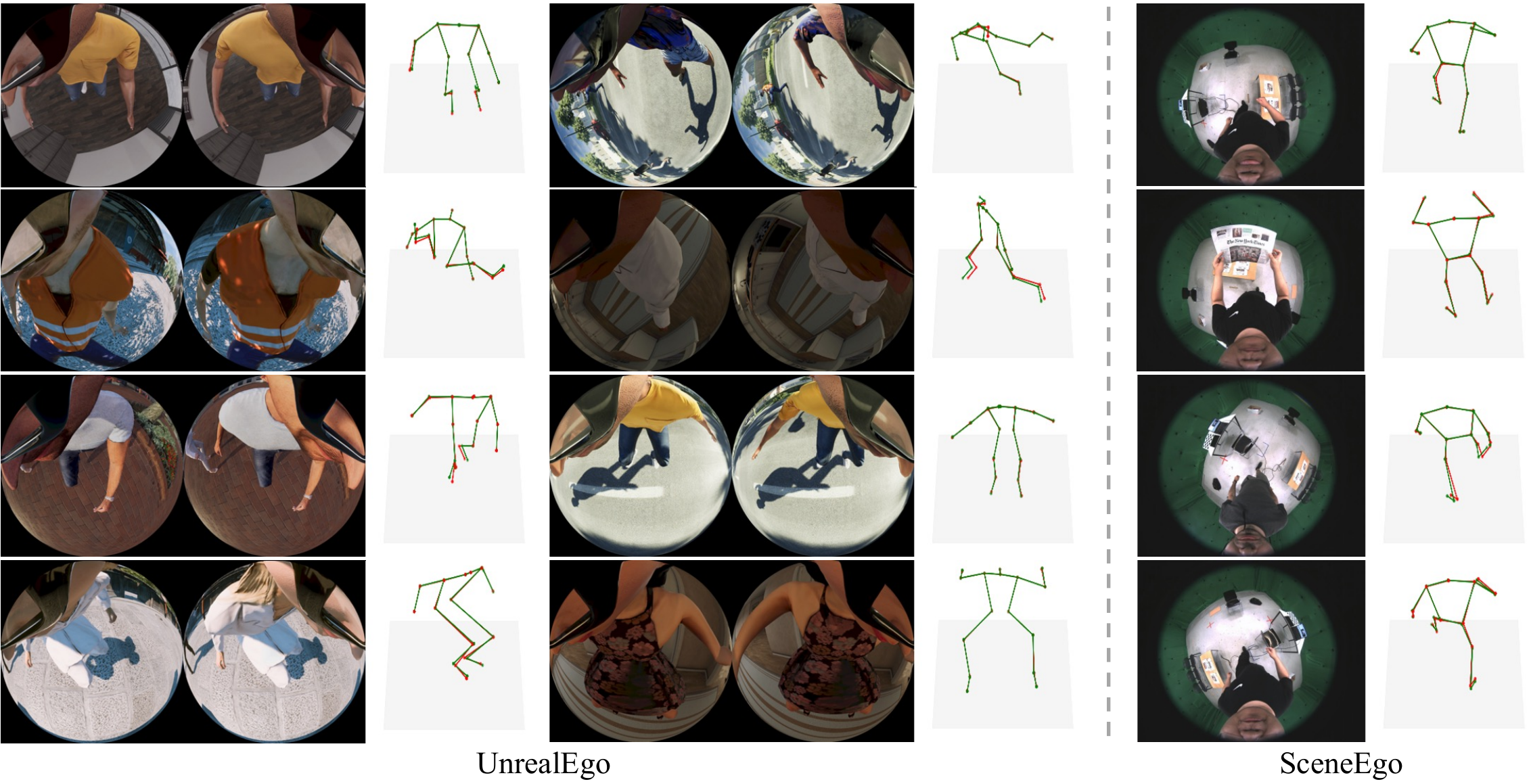}
    \caption{\small Qualitative visualization on UnrealEgo and SceneEgo. Ground-truths are colored in \textcolor{VisGreen}{green} and predictions are colored in \textcolor{red}{red}.}
    \label{fig:qualitative}
\end{figure*}

\noindent\textbf{Comparison with the state-of-the-art.}
In~\cref{tab:unrealego}, we compare our \ourmethod~with previous state-of-the-art methods on the UnrealEgo~\cite{hakada2022unrealego} dataset. We report both average and action-specific metrics. The results of competing methods are taken from their corresponding papers~\cite{hakada2022unrealego,kang2023ego3dpose}. As shown by the results, our approach achieves a significant advantage over all competing methods across all actions. Specifically, our method is 27.4mm better in the averaged MPJPE than the previous state-of-the-art Ego3DPose~\cite{kang2023ego3dpose}, pushing the best performance on this benchmark by 45\% lower error. In~\cref{tbl:sceneEgo}, we compare our method with previous egocentric body pose estimation approaches on the monocular SceneEgo~\cite{wang2023scene} dataset. Our method achieved an MPJPE that is 25.5mm lower than the previous best method. Note that SceneEgo~\cite{wang2023scene} requires ground-truth depth and semantic masks to train their model, but our method does not require them. Also, the previous best model was pre-trained on a large pose estimation dataset and fine-tuned on the SceneEgo dataset~\cite{wang2023scene}, but we only used the SceneEgo dataset itself to train our model. In~\cref{fig:qualitative},  we show qualitative results produce by our method on both datasets, which validates that \ourmethod~can generalize well to challenging poses. In summary, these results validate the effectiveness of our method on both multi-view and monocular egocentric pose estimation. \\

\noindent\textbf{Computation efficiency.}
In~\cref{tbl:flops}, we compare the computational efficiency of the proposed \ourmethod~with other competing models on the UnrealEgo and SceneEgo datasets, where we report the model parameters together with the number of FLOPs. The results show that our method is highly effective yet efficient. Specifically, on the UnrealEgo dataset, our method decreases the error over the previous best method by 45\% with only 7.9\% of the parameters and 13.1\% of the FLOPs. Similarly, on the SceneEgo dataset, our method reduces the MPJPE by 25.5mm compared to the previous best method, but requires only 60.7\% of the parameters and 36.4\% of the FLOPs.

\subsection{Ablation Studies and Discussions}
\label{exp:ablation}

\begin{table}[t]
    \begin{minipage}[b]{0.44\textwidth}
     \centering
    \caption{\footnotesize Ablation Study of attention operations on the UnrealEgo dataset. (DS-Attn denotes our \dsattn.)}
    \label{tbl:attnAblation}
    \begin{adjustbox}{max width=\textwidth}
    \begin{tabular}{cc|cc|cc}
    \toprule
    DS & Self & \multicolumn{2}{c|}{First Stage} & \multicolumn{2}{c}{Second Stage} \\
     Attn & Attn  & MPJPE & PA-MPJPE & MPJPE & PA-MPJPE \\
    \midrule
     & & 46.7 & 39.0 & - & - \\
     & \checkmark & 46.9 & 38.6 & 45.9 & 38.0  \\
     \checkmark & & 46.5 & 39.0 & 38.1 & 36.2  \\
    \checkmark  & \checkmark & 45.5 & 38.3 & 33.4 & 32.7  \\
    \bottomrule
    \end{tabular}
    \end{adjustbox}
    \end{minipage}%
    \hfill
    \begin{minipage}[b]{0.54\textwidth}
    \centering
    \caption{\footnotesize Ablation study of the stereo and monocular settings on the UnrealEgo dataset. Note here the feature extractors are also pre-trained using the corresponding settings.}
    \label{tbl:monocularStereo}
    \begin{adjustbox}{max width=\textwidth}
    \begin{tabular}{l@{\hskip 10pt}|cc|cc}
    \toprule
    & \multicolumn{2}{c|}{First Stage} & \multicolumn{2}{c}{Second Stage} \\
     Model Input & MPJPE & PA-MPJPE & MPJPE & PA-MPJPE\\
    \midrule
     Monocular Left  & 56.0 & 49.1 & 48.3 & 47.2 \\
     Monocular Right  & 56.3 & 49.0 & 48.7 & 47.3 \\
     Monocular Left + Right & 55.6 & 48.8 & 48.2 & 47.2 \\
    \midrule
    Stereo & 45.5 & 38.3  & 33.4 & 32.7 \\
    \bottomrule
    \end{tabular}
    \end{adjustbox}
    \end{minipage}
\end{table}

\begin{figure}[!t]
    \centering
    \includegraphics[width=\linewidth]{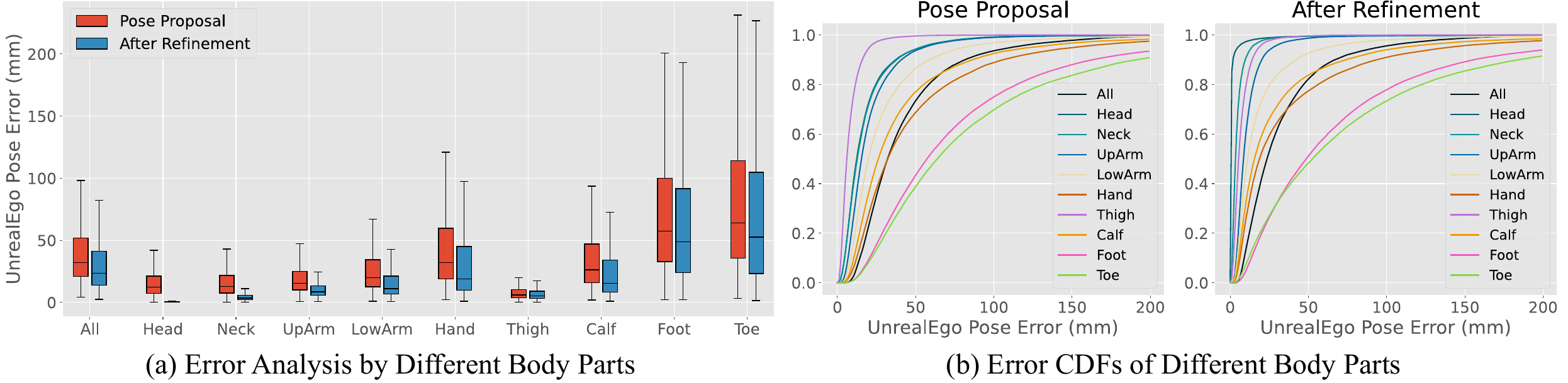}
    \caption{\footnotesize Error analysis of the pose proposal and the refined body pose estimation.}
    \label{fig:perJoint}
\end{figure}

\begin{figure}[!t]
\centering
    \includegraphics[width=0.9\linewidth]{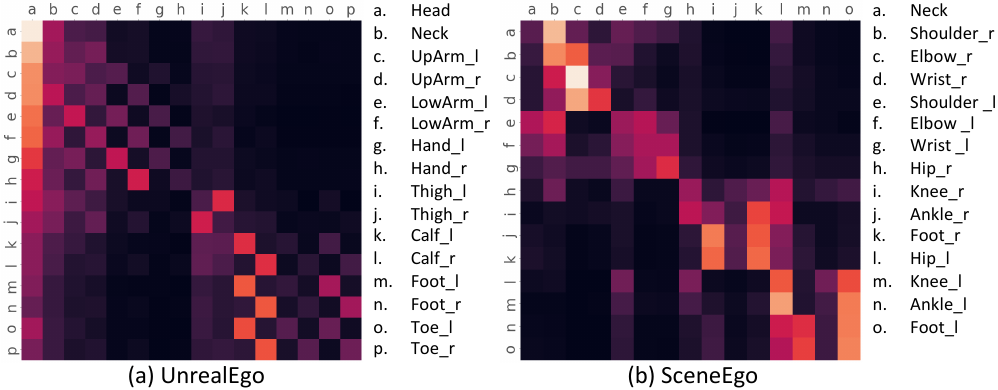}
    \caption{\small Heatmaps of averaged self-attention weights between different joints on the UnrealEgo and SceneEgo datasets.}
    \label{fig:selfAttn}
\end{figure}

\noindent\textbf{\rFormer.}
We first conduct experiments on the UnrealEgo dataset to study the two attention modules in our \rFormer. We present the results in~\cref{tbl:attnAblation}. We start with a baseline that only contains the \ppn~with the \rFormer~part being removed. Surprisingly, this simple baseline achieves a 14.1mm MPJPE advantage over the previous state-of-the-art. We speculate such an advantage is achieved by the usage of visual features. We then added \rFormer~but with only the self-attention operation, from which the refined pose estimation is only marginally better than the pose proposal. Subsequently, we add another version of \rFormer~that only contains \dsattn, this time the refined pose estimation improves MPJPE by 8.4mm over the initial pose proposal, validating its effectiveness in exploiting multi-view stereo features. Lastly, we add the full version of \rFormer~with both types of attention, from which the refined pose estimation is 12.1mm more accurate than the initial pose proposal. 
We therefore conclude that it is essential to combine the \dsattn~with self-attention, in which the JQTs interact with each other to propagate human kinematic information. 
To further understand how \rFormer~refines the pose proposal, in~\cref{fig:perJoint} we report the MPJPE improvement w.r.t. different body parts. Finally, we provide qualitative examples to better understand how the attention mechanisms work: In~\cref{fig:selfAttn}, we visualize the averaged attention weights between different joints in the self-attention operation on both datasets, from which we can see even if we do not explicitly introduce any kinematic-related losses in our model, the attention can implicitly infer the kinematic structures of a human body. \\

\begin{table*}[t!]
   \footnotesize 
  \centering
   \caption{\footnotesize  Error of different body parts when they are captured by different numbers of views on UnrealEgo. We report our reproduced result for the UnrealEgo model.}
  \label{tab:inOutFOV}
  \resizebox{\textwidth}{!}{
  \begin{tabular}{l|@{\hskip 8pt}c@{\hskip 8pt}|@{\hskip 8pt}c@{\hskip 8pt}c@{\hskip 8pt}c@{\hskip 8pt}c@{\hskip 8pt}c@{\hskip 8pt}c@{\hskip 8pt}c@{\hskip 8pt}c@{\hskip 8pt}c@{\hskip 8pt}|@{\hskip 8pt}c}
    \toprule
    \multirow{2}{*}{Method} & \multirow{2}{*}{Views}  & \multirow{2}{*}{Head} & \multirow{2}{*}{Neck} & Upper & Lower & \multirow{2}{*}{Hand} & \multirow{2}{*}{Thigh} & \multirow{2}{*}{Calf} & \multirow{2}{*}{Foot} & \multirow{2}{*}{Toe} & \multirow{2}{*}{MPJPE} \\
     &  & &  & Arm & Arm & & & & & \\
    \midrule
    \multirow{3}{*}{UnrealEgo} & 0 & - & - & - & 114.7 & 139.2 & 21.5 & 236.2& 282.3 & 319.1 & \multirow{3}{*}{80.1} \\ 
    & 1 & - & - & - & 104.8& 119.3& 39.5 & 241.9& 234.1& 292.7 &  \\ 
    & 2 & 49.1 & 45.4& 50.0& 74.3& 98.3& 16.3& 85.7& 122.1& 139.6 & \\ 
    \midrule
    \multirow{3}{*}{Ours} & 0 & - & - & - & 55.4 & 111.8& 19.1 & 147.2& 178.4& 179.3 & \multirow{3}{*}{33.4}   \\ 
    & 1 & - & - & - & 47.4 & 83.8 & 32.6 & 74.3& 119.9& 170.3 & \\ 
    & 2 & 1.73 & 5.3 & 11.4 & 18.6 & 31.7 & 7.2 & 31.0 & 71.8 & 81.5 & \\ 
    \bottomrule
  \end{tabular}}

\end{table*}

\noindent\textbf{Monocular v.s. Multi-view.}
We design \dsattn~to effectively infer the multi-view stereo information, which is crucial to accurate 3D localization. Here, we examine if the performance improvement is brought by the transformer model itself, or is a result of the effective stereo reasoning. To this end, we test our model on the UnrealEgo dataset in both monocular and stereo settings. The results are reported in~\cref{tbl:monocularStereo}. For the monocular setting, we treat every image as an independent sample. We tried three monocular settings: training and testing the model with the left view only, the right view only, and a mixing of both views. We got similar numbers for all three settings. The results show using monocular images decreased the performance of both \ppn~and \rFormer. Specifically, the monocular \ppn s~increases MPJPE by 10mm compared to the stereo variant. In addition, the monocular \rFormer s can only improve the MPJPE of pose proposal by 7.5mm, but in the multi-view setting \rFormer~can improve the MPJPE of the pose proposal by 12mm. From the results, we conclude: \textit{1) Multi-view input is important to the performance of both \ppn~and \rFormer. 2) While \rFormer~can refine the pose proposal with monocular inputs, it becomes more effective with multi-view inputs.} To further validate the conclusions, in~\cref{tab:inOutFOV} we presents error metrics for different body parts when they are captured by varying numbers of views.  We also include results for the heatmap-based UnrealEgo~\cite{hakada2022unrealego} for comparasion. From the findings, a clear correlation emerges between the localization accuracy of a joint and the number of views in which it is visible. When a joint is concurrently visible in both views, our method consistently achieves highly accurate localization.\\

\noindent\textbf{Error Analysis.}
We first study the error distributions of our method. We report the error distribution of different body parts in~\cref{fig:perJoint} (a), encompassing both pre-refinement and post-refinement results. Our observations reveal that, for the majority of joints, our \rFormer~exhibits a substantial capacity to reduce the errors in the pose proposals. Notably, except for the challenging \textit{Foot} and \textit{Toe}, the median error in refined poses even falls below the quartile error of the pose proposals, validating the effectiveness of our proposed \rFormer~in enhancing the overall accuracy of the pose estimation. On the other hand, it is worth noting that the \textit{Foot} and \textit{Toe} joints, although improved in the post-refinement phase, still exhibit considerable inaccuracy, limiting the overall accuracy of the pose estimation system. Furthermore, following~\cite{kang2023ego3dpose}, we visualize the error distributions in the form of Cumulative Distribution Function (CDF) for different body parts. The results are shown in~\cref{fig:perJoint} (b), including both the pre-refinement and post-refinement CDFs. Similar to the earlier observations, the accuracy of pose estimation is substantially improved after refinement. For example, for the \textit{Hand} joint, 68\% of the proposals have an error less than 50mm, and this number rises to 77\% for refined estimations under the same threshold. \\

\noindent\textbf{\rFormer~layers.}
In~\cref{tbl:numStages}, we report \ourmethod's performance with different numbers of layers of \rFormer, in which the JQTs iteratively interact with each other and the image features to refine the pose proposal. The results show that more layers of \rFormer~can lead to better accuracy of pose estimation. However, when the number of layers is greater than 3, the additional layers can only bring marginal improvement. For example, increasing layers from three to four only brings 0.4mm MPJPE improvement. Therefore, we adopt a 3-layer \rFormer~ in our model. \\ 

\begin{table}[t]
    \begin{minipage}[b]{0.46\textwidth}
     \centering
    \caption{\footnotesize Ablation study about how the number of refinement layers in \rFormer~affects the model performance on the UnrealEgo dataset.}
    \label{tbl:numStages}
    \begin{adjustbox}{max width=\textwidth}
    \begin{tabular}{l|c@{\hskip 5pt}c@{\hskip 5pt}c@{\hskip 5pt}c@{\hskip 5pt}c@{\hskip 5pt}c}
    \toprule
     Layers & 0 & 1 & 2 & 3 & 4 & 5 \\
    \midrule
     MPJPE & 46.7  & 37.9  & 35.1 & 33.4  & 33.0 & 32.8 \\
    \bottomrule
    \end{tabular}
    \end{adjustbox}
    \end{minipage}%
    \hfill
    \begin{minipage}[b]{0.5\textwidth}
    \centering
    \caption{\footnotesize Comparison of MPJPE (PA-MPJPE) between \ourmethod~and the heatmap-based UnrealEgo using different backbones on the UnrealEgo dataset.}
    \label{tbl:backbone}
    \begin{adjustbox}{max width=\textwidth}
    \begin{tabular}{l|cccc}
    \toprule
     Method & ResNet-18 & ResNet-34 & ResNet-50 & ResNet-101 \\
    \midrule
     UnrealEgo~\cite{hakada2022unrealego} & 79.1 (59.2) & 80.5 (60.1) & 80.1 (60.1) & 80.1 (60.5) \\
     Ours & 33.4 (32.7) & 29.5 (29.0) & 28.5 (28.1) & 27.9 (27.4) \\
    \bottomrule
    \end{tabular}
    \end{adjustbox}
    \end{minipage}
\end{table}

\begin{table}[t]
    \begin{minipage}[b]{0.44\textwidth}
     \centering
    \caption{\footnotesize Comparison between our approach and the baseline UnrealEgo method on the UnrealEgo dataset using different training strategies. $\dagger$ denotes our reproduced result. }
    \label{tbl:trainingStrategy}
    \begin{adjustbox}{max width=\textwidth}
    \begin{tabular}{l|c|cc}
    \toprule
      \multirow{2}{*}{Method}  & Training & \multirow{2}{*}{MPJPE} & \multirow{2}{*}{PA-MPJPE} \\
       & Strategy & & \\
    \midrule
     \multirow{3}{*}{UnrealEgo} & UnrealEgo & 79.1 & 59.2 \\
                    & UnrealEgo$^{\dagger}$ & 80.1 & 57.9\\
                    & Ours & 70.5 & 55.3 \\ 
    \midrule
    Ours & Ours & 33.4 & 32.7 \\ 
    \bottomrule
    \end{tabular}
    \end{adjustbox}
    \end{minipage}%
    \hfill
    \begin{minipage}[b]{0.53\textwidth}
    \centering
    \caption{\footnotesize Detailed ablation study on training strategies of PPN on the UnrealEgo dataset. Note the \ppn~performance is a bit better than the result in Tab.~\ref{tbl:attnAblation} because we used a newly pre-trained backbone.}
    \label{tab:ppnTraining}
    \begin{adjustbox}{max width=\textwidth}
    \begin{tabular}{c|c|c|c|c|cc}
    \toprule
     Feature & \multirow{2}{*}{OPT} & \multirow{2}{*}{ACT} & \multirow{2}{*}{\# Layer} & Heatmap & \multirow{2}{*}{MPJPE} & \multirow{2}{*}{PA-MPJPE}  \\
     Extractor &  &  &  & Pretrain &  &  \\
     \midrule
     UnrealEgo  & AdamW & GELU & 2 & \xmark & 48.3 & 40.4 \\
     \midrule
     \multirow{5}{*}{Ours} & Adam & Leaky & 3 & \xmark & 168.9 & 125.5 \\
     & AdamW & Leaky & 3 & \xmark & 59.6 & 48.5\\
     & AdamW & Leaky & 3 & \checkmark & 43.5 & 36.5 \\
     & AdamW & GELU & 3 & \checkmark & 42.6 & 36.1  \\
     & AdamW & GELU & 2 & \checkmark & 43.1 & 36.3 \\
     \bottomrule
    \end{tabular}
    \end{adjustbox}
    \end{minipage}
\end{table}

\noindent\textbf{Scaling up experiments.}
In~\cref{tbl:backbone}, we show our approach has a good scaling-up ability when using larger backbones. Specifically, we test our methods using different ResNet~\cite{he2016deep} variants on the UnrealEgo dataset. We keep the training recipe for all backbones the same, as described in~\cref{exp:settings}. We observe that the performance of our method gets consistently improves when we use larger backbones, which shows that our method scales well. In addition, we also report the performance of the previous heatmap-based UnrealEgo~\cite{hakada2022unrealego} method, which exhibits poor scaling-up ability when equipped with larger backbones. As we explained in~\cref{sec:intro}, this may be caused by the idling of visual features. \\ %

\noindent\textbf{Training Strategy.}
One surprising result of our paper is that even our  \ppnFull~excels the prior state-of-the-art method. Therefore, we raise a pertinent question: \textit{Is the superiority of \ourmethod~a consequence of a better training strategy instead of the model itself?} Notably, the most distinct difference in our training strategy, as compared to previous approaches~\cite{hakada2022unrealego,kang2023ego3dpose}, is that we followed the common practices~\cite{touvron2021training,dosovitskiy2020image,carion2020end} of transformer training to use the advanced AdamW~\cite{loshchilov2018decoupled} optimizer. In contrast, prior works employed the Adam optimizer~\cite{kingma2014adam}. To investigate this, we present the results of the UnrealEgo~\cite{hakada2022unrealego} baseline method using different training strategies in~\cref{tbl:trainingStrategy}. The outcomes clearly demonstrate that switching to our training strategy can indeed improve the MPJPE of the baseline method by around 9mm. Nevertheless, it is noteworthy that even with this improved baseline, our \ourmethod~still significantly outperforms it, validating the effectiveness of our method. In~\cref{tab:ppnTraining}, we report extensive ablation studies about how the training techniques affect the performance \ppn~when the \rFormer~part is removed. Here we investigate several factors: 1) the feature extractor architecture, 2) optimizer, 3) activation function, 4) number of MLP layers, 5) whether the feature extractors are pre-trained. The result shows the AdamW optimizer and the heatmap pre-training are the two most important reasons for \ppn's good performance.

\section{Conclusion}

We introduce \ourmethod, a new transformer-based egocentric pose estimation method. Our two-stage method first infers each joint's coarse from the global features, then uses a DETR-style transformer \rFormerFull~to refine the coarse locations by exploiting fine-grained stereo features and human kinematic information. Furthermore, we design \dsattn~to better exploit the multi-view stereo information. Our method achieves state-of-the-art with significant advantages over previous arts on two pose estimation datasets, including stereo and monocular settings. We hope our model can serve as a strong baseline approach for future research in this field.

\bibliographystyle{splncs04}
\bibliography{egbib}

\appendix

\section{Extra experiments}

\begin{wraptable}{R}{0.5\textwidth}
\centering
\vspace{-12mm}
\caption{\footnotesize Ablation study on 2D heatmap pre-training on the UnrealEgo dataset.}
\label{tbl:pretrain}
  \begin{adjustbox}{max width=0.5\textwidth}
    \begin{tabular}{l|c|cc|cc}
    \toprule
     & & \multicolumn{2}{c|}{First Stage} & \multicolumn{2}{c}{Second Stage} \\
     Dataset & Pretrain & MPJPE & PA-MPJPE & MPJPE & PA-MPJPE \\
    \midrule
     \multirow{2}{*}{UnrealEgo} &  & 48.8 & 40.1 & 36.5 & 35.1 \\
                    & \checkmark & 45.5 & 38.3 & 33.4 & 32.7  \\
    \midrule
    \multirow{2}{*}{SceneEgo} & & 182.5 & 119.6 & 122.9 & 97.2  \\
        & \checkmark & 120.3 & 87.9 & 93.0 & 74.3  \\
    \bottomrule
    \end{tabular}
\end{adjustbox}
\vspace{-5mm}
\end{wraptable}

\subsection{2D heatmap pre-training.}
Intuitively, 2D heatmap pre-training should tell the model what the appearance of a joint should look like, thus it can guide the deformable stereo attention to attend to relevant features, which can further help with accurately estimating the joints' 3D locations. In~\cref{tbl:pretrain}, we report how this pre-training influences the model performance on both the stereo UnrealEgo and the monocular SceneEgo datasets. On UnrealEgo, our pre-training improves the MPJPE of the pose proposal by 3.3mm and the final prediction by 3.1mm. However, on the SceneEgo dataset, such improvement becomes more significant, with 62.2mm for the pose proposal and 29.9mm for the final prediction. The reason is that as the SceneEgo dataset does not have stereo information, the appearance features become more important in localizing a joint. This experiment validates the effectiveness of our pre-training strategy.

\begin{figure*}[!ht]
    \centering
    \includegraphics[width=1.0\textwidth]{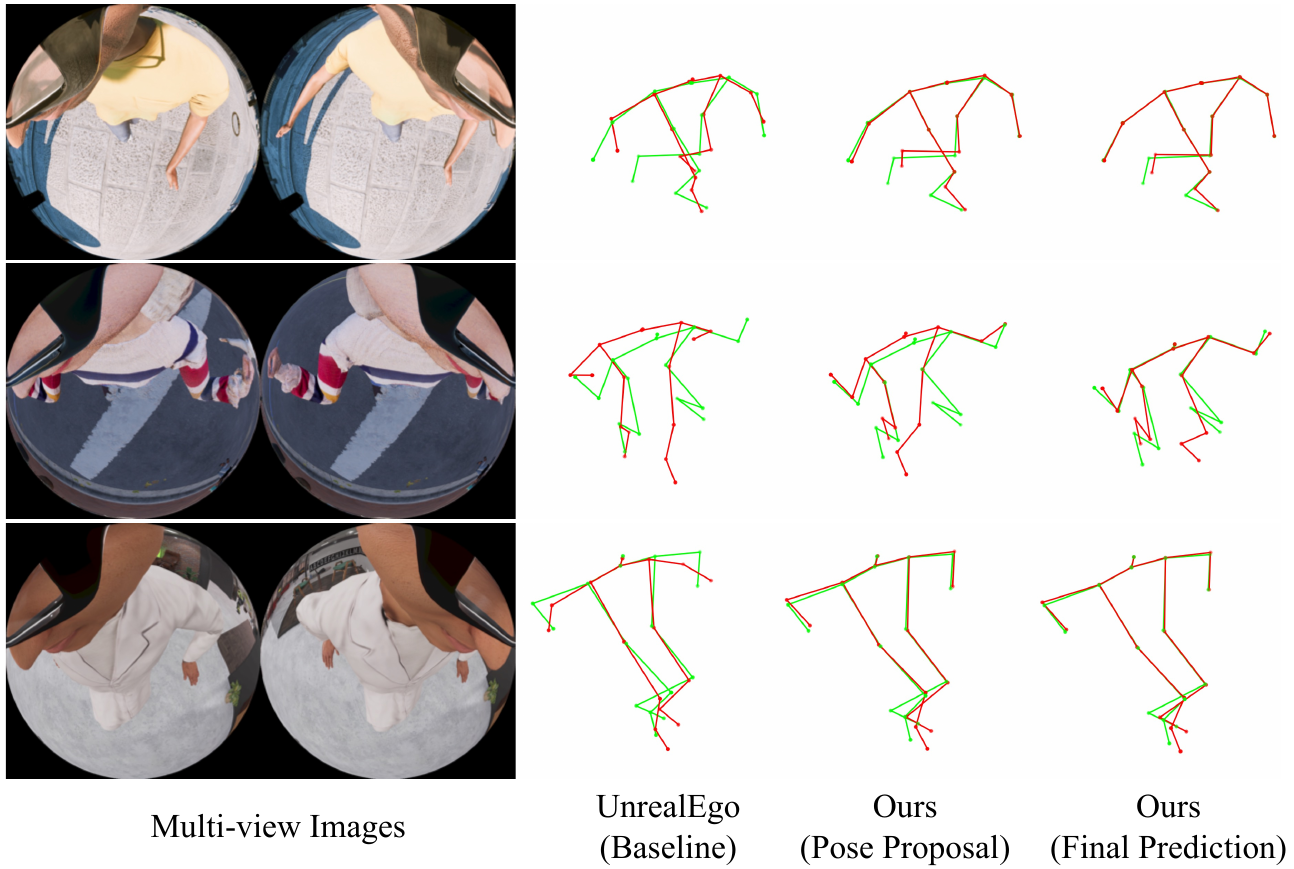}
    \caption{\footnotesize Qualitative comparison between EgoPoseFormer and the UnrealEgo baseline model on the UnrealEgo dataset. Ground-truths are colored in \textcolor{VisGreen}{green} and predictions are colored in \textcolor{red}{red}.}
    \label{fig:qualiCompare}
\end{figure*}

\subsection{Qualitative comparison with the baseline}
In Fig.~\ref{fig:qualiCompare}, we compare the qualitative failure cases of PPN (pose proposal), PRFormer (final prediction), and the baseline `UnrealEgo' model. It shows that most errors in our model's final prediction are caused by the joint invisibility problem (mostly in lower-body). Compared with PRFormer, PPN's estimations are more inaccurate because they are computed using the coarse global features. On the other hand, the baseline's performance is far from satisfactory even when the joints are captured by both cameras.

\begin{figure*}[!ht]
    \centering
    \includegraphics[width=1.0\textwidth]{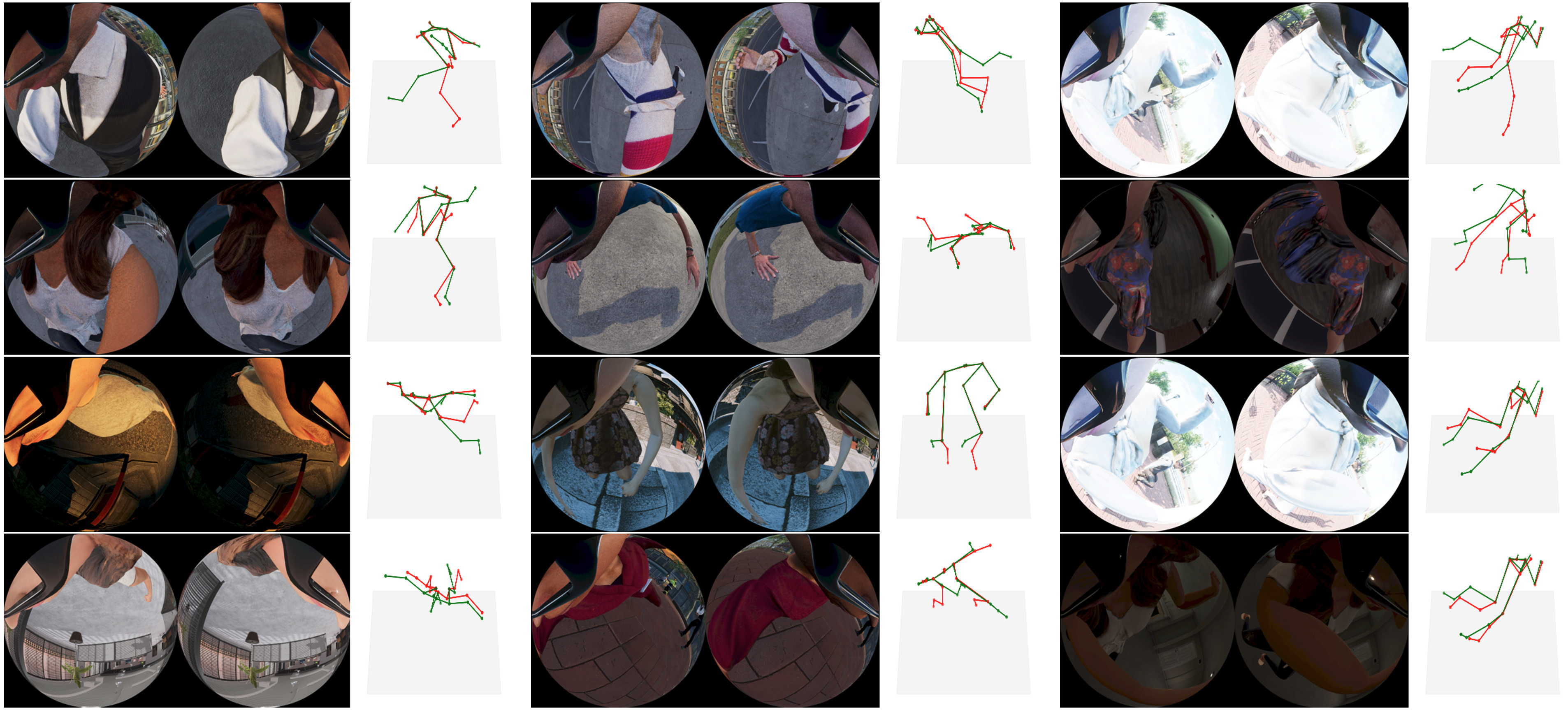}
    \caption{\footnotesize Visualization of \ourmethod's most inaccurate failure cases on the UnrealEgo dataset. Ground-truths are colored in \textcolor{VisGreen}{green} and predictions are colored in \textcolor{red}{red}.}
    \label{fig:fail}
\end{figure*}

\subsection{Qualitative failure cases}
In order to gain a more intuitive insight into the failure cases of our method, we present visualizations of some of the most inaccurate results in Fig.~\ref{fig:fail}. A clear observation is that the majority of these failure cases are attributed to the problem of joint invisibility. For instance, in the first example in the first column, the lower body of the wearer is entirely occluded by his upper body, leading to a substantial discrepancy in the estimated locations of the lower body joints. In another instance, illustrated in the last example of the first column, nearly half of the wearer's body extends outside the FOV of both cameras, causing severe inaccuracy for the estimated 3D pose. These examples underscore the impact of joint invisibility in egocentric 3D pose estimation. Although our method can indeed estimate the locations of invisible joints in some cases, as we explained in the main paper, such an estimation is achieved by jointly looking at the visible joints and the background scene. However, when a large part of the wearer's body is invisible, the estimated pose is still far from accurate. Therefore, achieving accurate localization for such joints remains an important and valuable topic for future research endeavors.

\subsection{Dependency of two stages}
\begin{wrapfigure}{R}{0.5\textwidth}
\vspace{-12mm}
\begin{center}
    \includegraphics[width=0.5\textwidth]{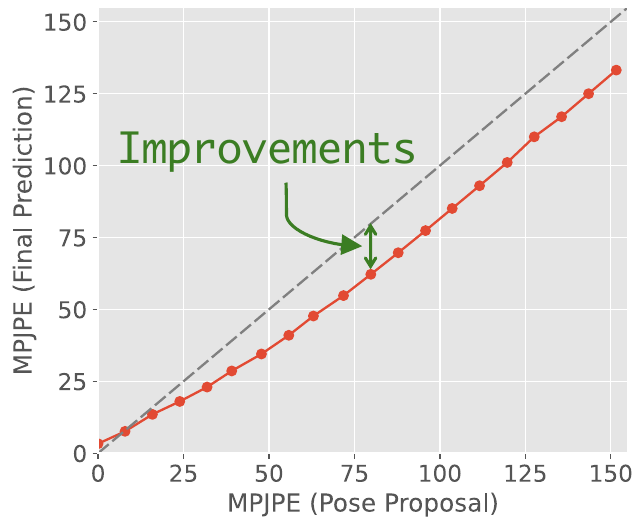}
\vspace{-8mm}
\caption{\footnotesize Dependency of the accuracy of PPN and PRFormer.}
\label{fig:dependency}
\end{center}
\vspace{-7mm}
\end{wrapfigure}

Here, we conduct an ablation experiment on the UnrealEgo dataset to check PRFormer's performance when the quality of the pose proposal varies. Specifically, we use perturbed ground truth, computed by adding Gaussian noises with different scales, to serve as pose proposals, based on which we use PRFormer to compute the refined pose estimation. We plot the dependency of the two stage's MPJPE in Fig.~\ref{fig:dependency} Left. The result suggests that although PRFormer's performance is positively related to the accuracy of the pose proposal, the refined pose estimation is always more accurate than the initial estimation, validating the effectiveness of our PRFormer.

\section{Potential ethical concerns}
Technically, one limitation of our PPN and PRFormer is that they assume headset wearers to have a full body, resulting in poor support for estimating body poses for individuals with disabilities, particularly those who have lost parts of their bodies. We acknowledge this limitation and plan to address it in future work. Another potential negative impact of our model is related to user privacy. For example, malicious agents could misuse the technology to analyze a user's body pose without their permission. The widespread use of such technology could also potentially lead to increased surveillance and tracking of individuals, raising ethical concerns about its use in public and private spaces. Another concern could be the risk of reinforcing biases present in the training data, which could lead to inaccuracies in pose estimation for certain demographic groups. Finally, there is the potential for dependency on this technology. For example, one may first need to buy a headset before using our model, which might reduce users' ability to perform tasks without it, affecting their autonomy and skill development.

\end{document}